Research Article

Amit Kumar Das*, Abdullah Al Asif, Anik Paul, and Md. Nur Hossain

# Bangla hate speech detection on social media using attention-based recurrent neural network

**Abstract:** Hate speech has spread more rapidly through the daily use of technology and, most notably, by sharing your opinions or feelings on social media in a negative aspect. Although numerous works have been carried out in detecting hate speeches in English, German, and other languages, very few works have been carried out in the context of the Bengali language. In contrast, millions of people communicate on social media in Bengali. The few existing works that have been carried out need improvements in both accuracy and interpretability. This article proposed encoder–decoder-based machine learning model, a popular tool in NLP, to classify user's Bengali comments from Facebook pages. A dataset of 7,425 Bengali comments, consisting of seven distinct categories of hate speeches, was used to train and evaluate our model. For extracting and encoding local features from the comments, 1D convolutional layers were used. Finally, the attention mechanism, LSTM, and GRU-based decoders have been used for predicting hate speech categories. Among the three encoder–decoder algorithms, attention-based decoder obtained the best accuracy (77%).

## 1 Introduction

Social media is a trendy and most importantly an easy way for people to publicly share their feelings and opinions and interact with others online. Nowadays, it has become a part of human life. It is a platform where people are easily harassed or targeted by others, expressing hate in sexism, racism, political, or any other forms. Cyber oppression, online nuisance, and blackmailing using these social sites are also increasing rapidly. According to the survey of Pew Research Center institute in 2017, among 4,248 US adults, it is found that 41% of Internet users have personally faced online harassment, and 66% have witnessed this harassment directed to others. Nearly one out of five Americans, i.e., 18% of users, have experienced severe harassment such as physical threats, stalking, or sexual harassment [1,2]. These types of incidents happen through the sharing of inappropriate images, offensive remarks, and messages. Identifying and removing such contents is a task carried by social media researchers' attention throughout the years. In recent years, along with the impact and interest in online hate speech detection on social media, this task's automation has steadily increased.

* **Corresponding author: Amit Kumar Das,** Computer Science & Engineering (CSE), East West University, Dhaka, Dhaka, Bangladesh, e-mail: amit.csedu@gmail.com
**Abdullah Al Asif, Anik Paul, Md. Nur Hossain:** Computer Science & Engineering (CSE), East West University, Dhaka, Dhaka, Bangladesh

For people's interaction, Facebook is the preferable medium and one of the most popular social networking sites worldwide, with 2.23 billion monthly active users [3]. According to BTRCs November 2017s report, it is found that there are approximately 25–30 million Facebook users in Bangladesh [4]. Among them, 72% are males, and 38% are females. Dhaka is considered to be second among all the world cities, with the largest number of active Facebook users [5]. Seventy-three percent of women and 49% of students in Bangladesh face cyberbullying online Bangladesh [6].

Bangla language is considered to be the seventh most spoken native language by the population in the world. Almost 205 million speakers speak Bangla as their native language, which is around 3.05% of the total world population [7,8]. There are approximately 81.7 million Internet users in Bangladesh. Among them, active social media users are 30 million, and 28 million people access social media using their mobile phones [5,9]. Forty-two million Facebook users use the Bangla language to comment, post, or share on social media, which is almost 1.9% of total Facebook users. In other social media sites, the Bangla language is used by users too, and the use of the Bangla language is increasing day by day in all social sites. Much research has been done on abusive text detection in the English language using social networks.

In this article, the following contributions are made:
- A dataset of 7,425 Bengali comments was created. The comments were collected from various Facebook pages using Facebook Graph API. Since the Graph API was limited for us, we have also collected the comments manually. These comments were then classified into seven categories: Hate speech, aggressive comment, religious hatred, ethnical attack, religious comment, political comment, and suicidal comment.
- A Bangla Emot Module was created that helps to detect the emotions lying behind the emoji and emoticons. This helps to clarify the type of hate speech more clearly for Bangla Language. The related details are discussed later.
- Supervised machine learning algorithms such as attention LSTM- and Gated Recurrent Unit (GRU) based decoders was applied to the model. An accuracy of 74% was obtained from the aforementioned algorithms. The model was then improved by using the Attention mechanism with an accuracy of 77%.

## 2 Related work

The research interest in online hate speech detection has increased over the past few years. This occurred due to the impact of sharing hatred and other emotions on social media. For this, the task of detecting hate speech automatically has grown significantly [10,11]. Identifying a message containing text as hate speech or not is usually an easy task. Different types of ML algorithms were applied in English Dataset to classify abusive contents on social media. In previous studies [12,13], the authors have used character n-grams, word n-grams, and word skip-grams as features in their SVM model. The model acquired an accuracy of 78% in identifying comments across three classes (hate, offensive, and normal). Two types of hate speech detection model (context-aware logistic regression model and a neural network model with learning components) have been used to increase the performance by around 4% in F1 score. Later, combining these two models improved the accuracy by another 7% in the F1 score [14]. C. Nobata, J. Tetreault, A. Thomas, Y. Mehdad, and Y. Chang developed an online ML approach to identifying hate speech and classified sentences into Hate Speech, Derogatory, and Profanity. To measure the different aspects of the user's comment, they used the Vowpal Wabbits regression model along with N-grams, Linguistic, Syntactic features [15]. E. Chandrasekharan, M. Samory, A. Srinivasan, and E. Gilbert used the Bag of Communities (BoC) approach as a new concept to identify abusive content from a vast online community. The author also compares Naive Bayes (NB), Linear SVC, Logistic Regression algorithms for text classification. Naive Bayes (NB) performs best among all in all conditions [16]. In previous study [17], the author has used four measures to identify hate speech in English. They have used Facebook data to determine the type of hate speech. Their first stage was the exploring stage. They have identified some pages in this step that produce hate speech comments. Not all the pages' posts contained hate speech, so they used a regular speech filter to separate hate speech. They have also performed a sentimental analysis. I. M. A. Niam, B. Irawan, C. Setianingsih, and B. P. Putra have used Latent semantic analysis in their paper, a prevalent natural language processing (NLP) method. To extract information from the image LSA method was used to extract data from the image. They have used Twitter to gather information. They have got an average accuracy of 57.9% [18]. K. Nugroho, E. Noersasongko, A. Z. Fanani, and R. S. Basuki have applied a random forest approach on a Twitter dataset of 14,509 to identify hate speech and offensive words. The method was then compared with the AdaBoost and Neural Network to obtain higher accuracy. Random forest performed the best among the three, with an accuracy of 0.722 [19].

Abusive messages and hateful speech are shared in languages other than English too. Researchers' tried to solve them also. Z. Mossie, J. H. Wang, and D. Nagamalai developed an apache spark-based model to classify Facebook posts and comments that are written in the Amharic Language. The comments were classified into two categories: hate and non-hate. The model was implemented with Random forest and Naive Bayes for learning, and feature selection Word2Vec and TF-IDF were used. The model based on word2vec embedding performed best with 79.83% accuracy and performed well with extensive data [20]. N. Aulia and I. Budi conducted several ML approaches to explore hate speech detection in long text Indonesian data. They have created a new Indonesian hate speech dataset from Facebook. Support Vector Machine (SVM) performs best as its classifier algorithm using char quad-gram, TF-IDF, word unigram, and lexicon features giving the f1-score of 85% [21]. In previous study [22], the author created a dataset named HASOC developed from Twitter and partially from Facebook. The HASOC track aims to improve performance in Hate Speech for English, German, and Hindi languages. For the classification of the dataset, three subtasks were performed. Sub-task A is a coarse-grained binary classification in which tweets are classified into two classes: Hate and Offensive (HOF) and Non-Hate and offensive. Hate Speech and offensive posts from the subtask A are further classified into three categories in subtask B: (HATE) Hate speech, (OFFN) Offensive and (PRFN) Profane. Only posts labeled as HOF in subtask A are included in subtask C. The two categories in subtask C are Targeted Insult (TIN) and Untargeted (UNT). LSTM networks processing word embedding input were most commonly used as an approach, and the best system's performance for identifying Hate Speech for English, Hindi, and German was a Macro-F1 score of 0.78, 0.81, and 0.61, respectively.

Related to the Bangla language, researchers are still researching for the proper method to give the best accuracy [23,24]. The latest work on Bangla hate speech detection was Hateful Speech Detection in Public Facebook Pages for the Bengali Language [25]. The author had developed ML algorithm-based model, as well as GRU-based deep neural network model for classifying users' comments on Facebook pages and annotated 5,126 Bengali comments and classified them into six classes: Hate Speech, Communal Attack, Inciteful, Religious Hatred, Political Comments, and Religious Comments [25].

In this article, 2,500 comments were collected (after cleaning 5,126 comments) from ref. [25]. The rest 4,925 comments were collected from commentators using Facebook Graph API from different pages under Facebook's privacy policy. This article classified the comments into seven classes, whereas, six classes remain the same as ref. [25] and the last one was later introduced as "Suicidal Comment." Nowadays, suicidal comments are increasing on social media, and it influences the social norms negatively [26]. So, it is one kind of hate speech.

S. C. Eshan and M. S. Hasan proposed a system based on Bengali Unicode text that implements both count vectors and TF-IDF for unigram, bi-gram, and tri-gram. Then, MNB and SVM (linear and RBF) classifiers were used [27]. This system only dealt with Unicode Bengali words while in another paper, they have dealt with both words and emoticons [28]. Besides MNB and SVM algorithms, they have implemented the Convolutional Neural Network (CNN) with Long Short-Term Memory (LSTM). SVM with a linear kernel performed best with 78% accuracy among three algorithms. S. C. Eshan and M. S. Hasan have worked with 2,500 Bengali data which were collected only from popular Facebook pages [27]. However, in both papers, the preprocessing technique, like stemming, was absent. In previous study [29], a preprocessing method like stemming was used with a classifier to get the root form of a Bengali word. But the preprocessing technique did not follow proper Bengali grammatical rules for which it could only remove a Bengali word's suffixes. So, getting better performance was not possible for all types of data.

For classifying user's comments on Facebook pages in ref. [25], the authors have developed ML algorithm-based model, and as well as GRU based deep neural network model. They have created a dataset of Bangla comments and classified them into six different classes (i.e., Political Comments, Hate Speech, Communal Attack, Inciteful, Religious Hatred, and Religious Comments). They were the first ones to contribute 5,126 comments to the field of hate speech detection. Several ML algorithms were tested and their performance was compared. Random Forest attained an accuracy of 52.20%. Among them, the GRU-based model performed better with an accuracy of 70.10%.

The researchers have applied several ML models based on algorithms for Bangla hate speech detection in the previous studies. In this article, the Attention-based Recurrent Neural Network model was performed in order to detect Bangla Language hate speech detection. For the first time, Attention mechanism is applied to the field of hate speech detection. A new module was also created for sentiment analysis using emoticons for Bangla Language.

## 3 Method

The model of our proposed work consists of few processes. Initially, data were extracted using Facebook API and were saved into a CSV file. Data were split into a training set (80%) and a testing set (20%). Label Encoder was employed on the training and testing set to convert them into the corresponding binary values to feed into ML approaches as input [27,30].

Data were preprocessed using some preprocessing methods, as shown in Figure 1 to extract information, and also, information was extracted from emoticons and emojis to detect the type of speech. Bangla natural language tokenization was used to split sentences into words. Features were extracted using TF-IDF vectorization, and word embedding was also used. And finally, classification approaches such as CNN, Bidirectional LSTM, and GRU were applied for comparing their performance. Further applying Recurrent neural network (RNN) with Attention Mechanism for text classification. Finally, the performances of all the classification approaches have been analyzed and compared.

There are two categories of abusive text classification: binary classification and multi-class classification. Through binary, only whether a message is abusive or not can be decided, where in multi-class classification can determine a line whether it is a hate speech, a group of people insulting a particular person, or anger, etc. For the English Language, binary classification is much easier than multi-class classification. Here, a manual algorithm was developed to detect abusive Bangla text for experimenting.

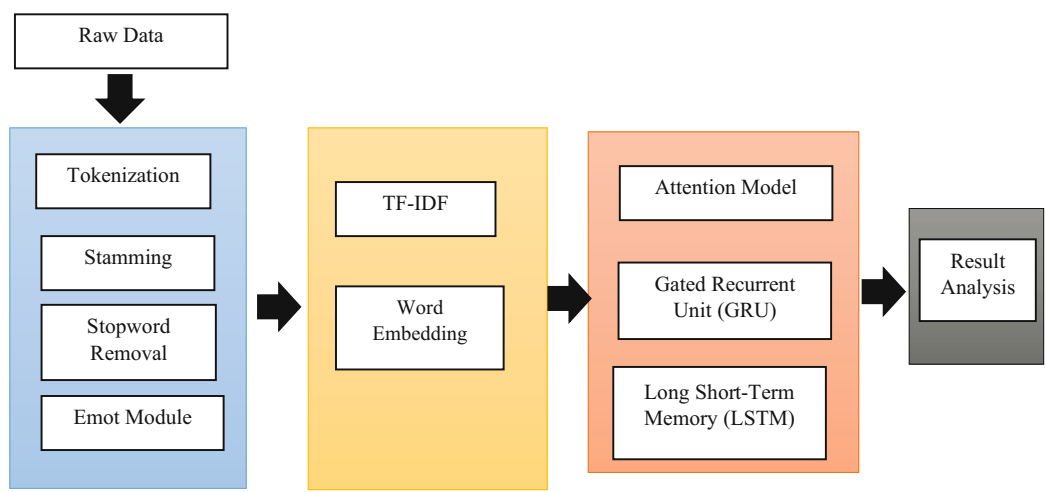

**Figure 1:** Infrastructure of our proposed model.

In this procedure, our work aims to discover a new way to detect hate speech. For this reason, the data were divided into seven categories: (a) Hate Speech, (b) Aggressive Comment, (c) Religious Hatred, (d) Ethnical Attack, (e) Religious, (f) Political Comment, and (g) Suicidal Comment.

Figures 2 and 3 show number of different types of hate speech and some of its examples along with the class type.

The popular public pages are the best options for collecting hate speech comments because of spreading toward individuals, cultures, and communities. In this study, public pages of celebrity, politician, model, actor, singer, news portal, and players were chosen because the hate spread through their hateful comments on such public pages. This type of page has millions of likes and followers, and by the posts and comments connect with lots of people every day. Initially, at least one page from these specifications was selected for the creation of data collection. The comments were collected from some most popular public pages in Bangladesh: ProthomAlo news, Bangladesh Pratidin and Ittefaq, Independent news,

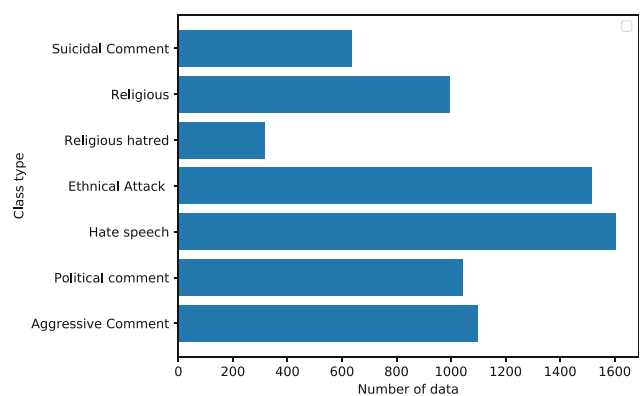

**Figure 2:** Categories of hate speech.

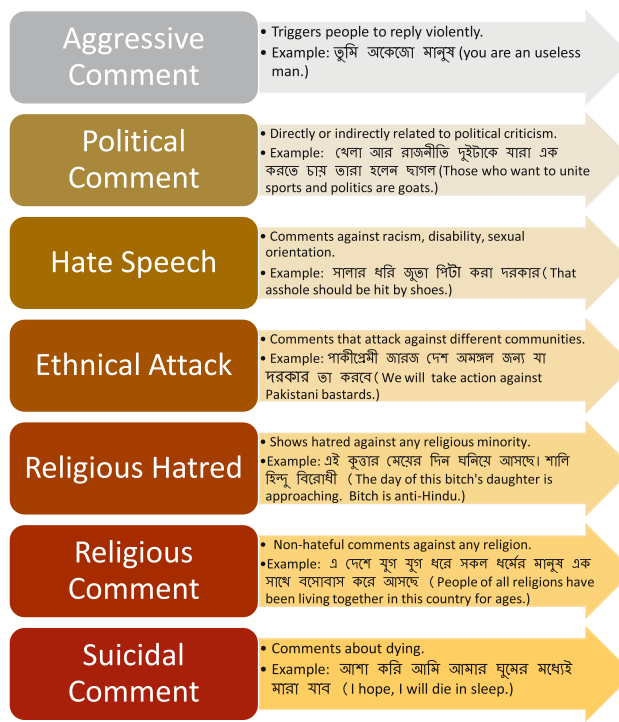

**Figure 3:** Displays the number of data each class contains.

Mashrafe Bin Mortaza, Shakib Al Hasan, SalmoN TheBrownFish, Tasinnation, Tahsan, Model Arif Khan, Naila Nayem, and some other models. Due to privacy protection, only public comments are collected without the information of the commenters. These kinds of pages belong to celebrities on Facebook, political pages of the ruling party's official page, and Bangladesh's famous cricketers. These pages have around 0.077, 0.15, 3.5, and 6 million followers, respectively, and are found every day with an average of 2 posts with 6k reactions and 300 comments per post on social media.

Among them, hateful comments are around 6,020, which are labeled as Hate speech, Aggressive comment, Religious hatred, and Ethnical attack. The hateful comments were divided into four classes, and the non-hateful classes are Religious Comment and Suicidal comment. Political Comment is considered to be the one with both hateful and non-hateful comments of politics. For a clear understanding of hateful speech, seven classes were created totally.

After creating a dataset from different Facebook pages, the raw text was processed to clean the punctuations, bad characters, stemming, etc. There are five parts to the preprocessing technique:

**Removal of bad characters, punctuations, etc.**: After the collection of raw data texts, the dataset has been cleaned by removing the non-letters such as the punctuations (comma (,), dot (.), semicolon (;), hyphen (-), underscore (_), exclamatory mark (!), question mark (?), etc.). Any kind of token was deleted, which has a frequency of less than 5. For a better result, the noisy and bad characters were removed to support the system. This was needed to be done so that the Unicode encoding works properly.

**Tokenization of string**: It is a process of splitting a sequence of strings into pieces such as phrases, words, symbols, etc. Figure 4 shows the splitting process from a sentence into words. These pieces are called tokens. Our next step of preprocessing was tokenizing the pure Bengali Text. Some characters such as punctuation marks were discarded during this process of tokenization.

Sample input: আমি ওর হাতগুলি ভেঙে দিয়েছিলাম
(I broke his arms)
Sample output: 'আমি', 'ওর', 'হাতগুলি', 'ভেঙে', 'দিয়েছিলাম'
('I', 'broke', 'his', 'arms')

**Figure 4:** Example of tokenization.

**Bangla Stemmer**: Stemming is the process of extracting the root word from the given word. The basic principle of stemming is to simplify the various types of grammar/word to their root, stem, or base form as shown in Figure 5. According to the rules, words can get inflected in any natural language. Bengali words are mainly inflected by verbal and nominal inflections [31,32]. In this study, the model is only dealing with a verb and noun inflection.

Sample input: 'আমি', 'ওর', 'হাতগুলি', 'ভেঙে', 'দিয়েছিলাম'
('I', 'broke', 'his', 'arms')
Sample output: 'আমি', 'ওর', 'হাত', 'ভেঙে', 'দেই'
('I', 'break', 'his', 'arm')

**Figure 5:** Example of Bangla stemmer.

**Removal of Bangla stopwords**: Stopwords are the commonly occurring collection of terms that do not contribute important details to text classification. In these steps, words that do not add much meaning to a Bengali sentence were removed. Figure 6 shows one of the examples of removing Bangla stop words. There is a stopword list available consisting of around 400 words [33]. Social media includes comments incorporated with misspelled words or short forms for which it would be difficult to identify the words. So that, frequently used stopwords with different formats of mis-spelling have been added to the list.

Sample input: 'আমি', 'ওর', 'হাত', 'ভেঙে', 'দেই'

('I', 'break', 'his', 'arm')

Sample output: 'হাত', 'ভেঙে', 'দেই'

('break', 'arm')

**Figure 6:** Example of removal bangla stopwords.

**Bangla Emot Module**: Emojis and emoticons are used in social media more often to express humans emotions. A Bangla Emot module was created to detect the feelings hidden behind this emojis and emoticons for Bangla Language. The meaning of each emojis and emoticons was taken from ref. [34], and their English meaning was converted to Bangla. This python module extracts information from emoticons and emojis, as shown in Figure 7, to detect the type of speech. As a result, it becomes easy to know what kind of emotions the user is sharing, and it will be easier to categorize people's comment.

Sample input: 'হাত', 'ভেঙে', 'দেই', '😠'

('break', 'arm', emoji)

Sample output: 'হাত', 'ভেঙে', 'দেই', 'ঘৃণা'

('break', 'arm', 'Anger')

**Figure 7:** Example of Bangla emot module.

Emoticon and emoji play an essential role in social media. People use both emojis and emoticons. There are some differences between them. Generally, an emoticon is a set of punctuation marks, letters, and numbers arranged to resemble a human face. For example, :-D means laughing or a big grin, :-O is for a surprise, etc. Emoji is a pictogram, a small picture that can show anything. The word emoji essentially means "picture-character" [35].

This article has dealt with both emoji and emoticon. All the emojis and emoticons are taken from a reliable source [36]. Each emoji has a Unicode that alludes to that particular emoticon [37]. The Unicode and punctuation marks are stored in emoticons and emoji data dictionary and give them a name what they mean. The names are as of now recorded in a site called Emojipedia [34]. So, when the function is run into the dataset it checks each sentence to find the Unicode or punctuation and convert it into Bangla text through the custom made "emot to Bangla text" dictionary.

Feature Selection Method is an essential part of the classification of tasks. TF-IDF is commonly chosen as a method for the extraction of features in various NLP tasks. The value TF-IDF represents the importance of a term in a sentence, paragraph, or document known as corpus. The frequency of a term $t$ occurring in document $d$ is determined by TF (term frequency). The detail about the importance of a term $t$ is provided by IDF (Inverse Document Frequency). The TF-IDF weight (of term $i$) for normalization is measured by:

$$w_i = (TF_i \times \log(N/n_i)) \bigg/ \sqrt{\sum_{i=1}^{n} (TF_i \times \log(N/n_i))^2}.$$

Normalize generally means forcing all values to fall within a specific range, usually between 0 and 1. The term weight has been normalized so that the longer documents are not unfairly given more weight.

Word embedding represents words in vector forms giving similar encoding for similar words in vector space. Figure 8 shows that vector representation is formed for a previous fixed sized vocabulary from a text's corpus using word embedding procedure. In the neural network, the embedding layer was used, which, along with the model in the training phase, learns the representations of words in vector form. After selecting the preprocessing steps and features, data and features were filled into the correct models using popular algorithms.

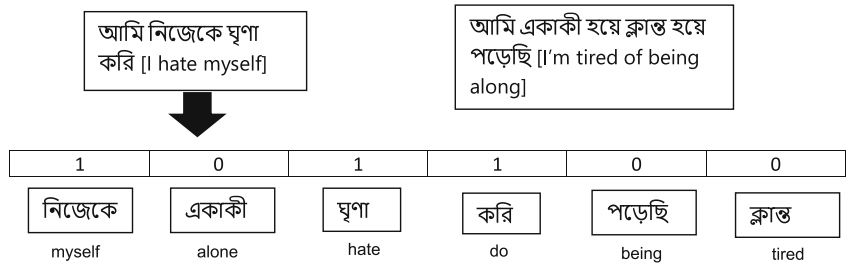

**Figure 8:** Example of word embedding.

This article proposed encoder–decoder-based architecture with a recurrent neural network. This method has become an effective and standard approach for both neural ML and sequence-to-sequence (seq2seq) prediction in general. An encoder is a network that takes the input and outputs a feature map/vector. These feature vectors hold the information, the features that represent the input. So, Convolutional Neural Network (Conv1D) has been used as an encoder to import the feature vector from the text data. On the other hand, the decoder is again a network (usually the same network structure as the encoder but in opposite orientation) that takes the feature vector from the encoder and gives the best closest match to the actual input. In this article, RNN-based decoder has been used such as LSTM, GRU, and Attention decoder to predict the final outcome. Finally, these models have compared each other and evaluate the best performance among them.

**Attention Mechanism**: A neural network is seen as a simplified attempt to imitate the human brain's actions by using the Additive Attention Mechanism in our experiment. Attention Mechanism is also a streamlined attempt that focuses on the few relevant things of the selective activities. In this article, learning the text's characteristics using a bidirectional RNN is necessary because the actual information of a text in a sentence is related to a text that is lying in front of it and to the text content lying behind it. Text representation is implemented from the learning phase using the bidirectional RNN method as shown. The two directions are then spiced together to learn the feature vector so that the semantics' eigen vector becomes more prosperous and more comprehensive relative to the unidirectional RNN.

At the same time, the attention mechanism was added to the network model so that each word learns its weight. Heavier weight for keywords and lighter weight for keywords for which important features becomes more noticeable.

The attention mechanism was destined to help retain long source sentences in neural machine interpretation. Instead of building a solitary context vector out of the encoder's last hidden state, the attention model comes to play to make easy routes between the context vector and the whole source input. The general structure of the model used in this article is given below.

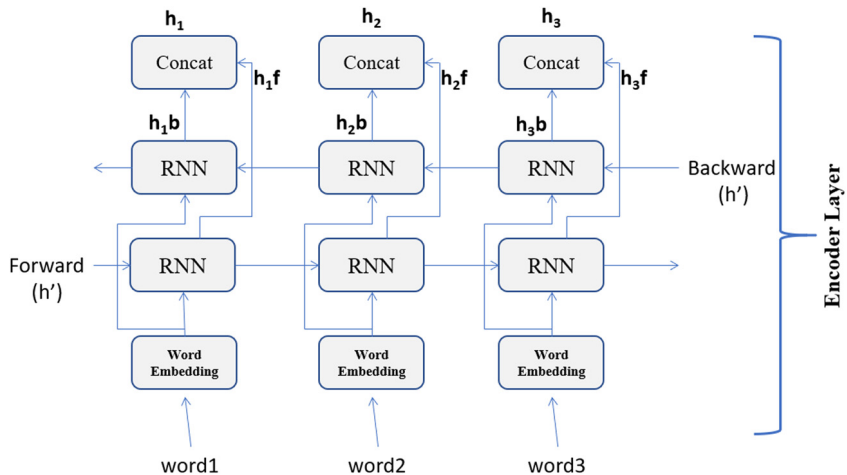

**Figure 9:** Encoder layer.

Figure 9 shows that every input sequence produces a hidden state while passing through the encoder layer. The encoder is a bidirectional RNN (LSTM) with forward (h) and backward hidden state (h'). Both states concatenate to produce the encoder state. This state is designed in a particular way both previous and subsequent words are included in a single word. Figure 10 represents the decoder layer and attention layer. The attention and decoder layers are explained below:

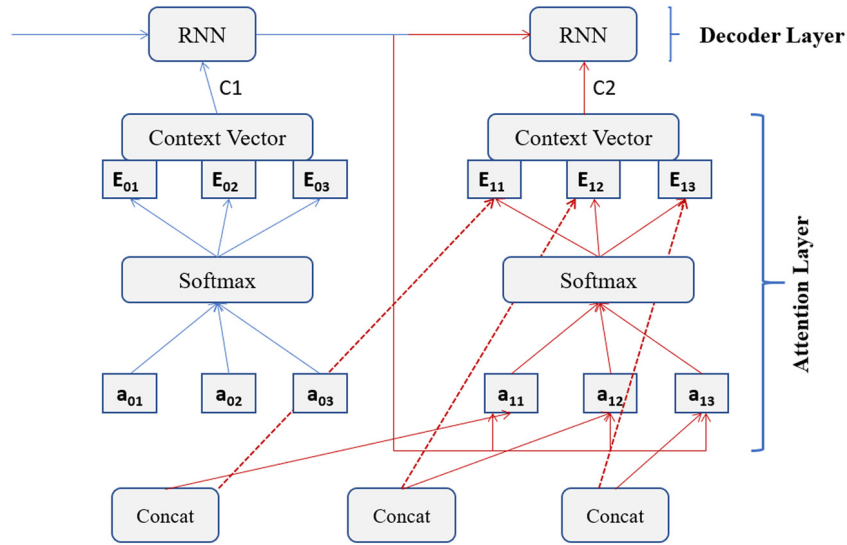

**Figure 10:** Decoder and attention layer.

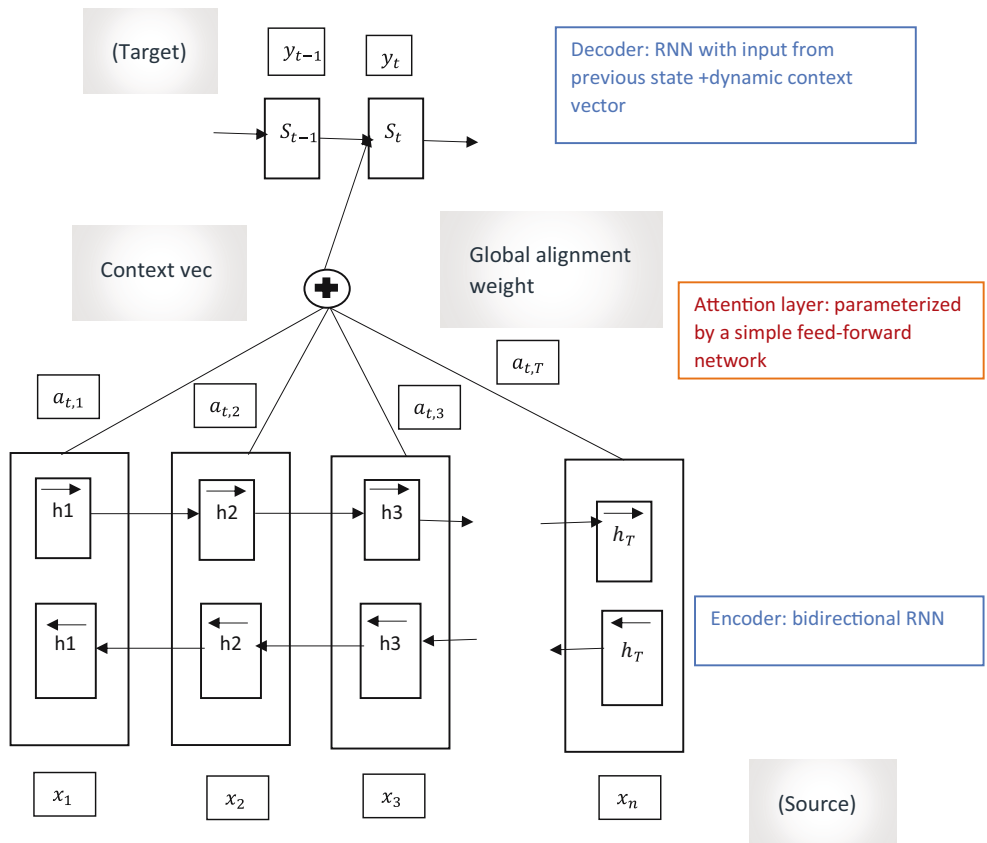

**Figure 11:** Additive attention mechanism.

In Figure 11, the decoder network has a hidden state $s_t = f(s_{t-1}, y_{t-1}, c_t)$ for the output word at position $t$, $t = 1, \ldots, m$ where the context vector $c_t$ is a sum of hidden states of the input sequence, weighted by alignment scores:

$$c_t = \sum_{i=1}^{n} \alpha_{t,i} h_i \quad \text{Context Vector for output } y_t$$

$$\alpha_{t,i} = \text{align}(y_t, x_i) \quad \text{How well two words } y_t \text{ and } x_i \text{ are alligned.}$$

$$= \frac{\exp(\text{score}(s_{t-1,h_i}))}{\sum_{i=1}^{n} \exp(\text{score}(s_{t-1}, h_i))} \quad \text{Softmax of some predefined alignment score.}$$

The alignment model assigns a score $t$, $i$ to the pair of input at position i and output at position $t$, $(y_t, x_i)$ based on how well they match. The set of $\alpha_t$, $i$ are weights defining how much of each source hidden state should be considered for each output. In Bahdanau's paper, the alignment score $\alpha$ is parameterized by a feed-forward network with a single hidden layer, and this network is jointly trained with other parts of the model. The score function is therefore in the following form, given that tanh is used as the non-linear activation function: Alignment function is $\text{score}(\mathbf{s}_t, \mathbf{h}_i) = \mathbf{v}_a^\top \tanh(\mathbf{W}_a[\mathbf{s}_t; \mathbf{h}_i])$ [38], where both $v_a$ and $w_a$ are weight matrices to be learned in the alignment model.

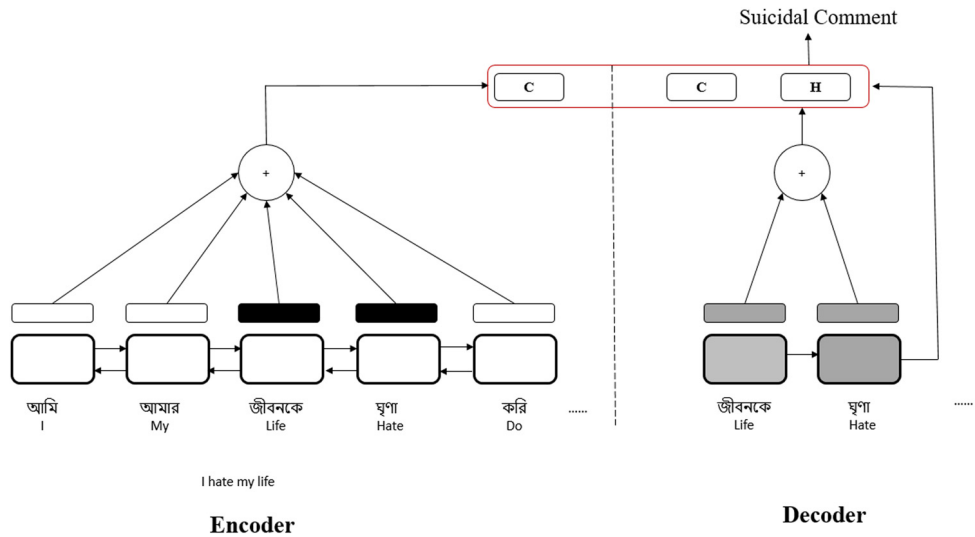

**Figure 12:** Illustration of the encoder and decoder attention functions combined in Bangla language.

In Figure 12, two context vectors (marked "C") are computed from attending over the encoder hidden states and decoder hidden states. Using these two context and current hidden states (H), a new word is generated and added to the output sequence and predicts the speech category.

## 4 Results and discussion

Various ML algorithms were applied to the collected dataset. The dataset is then preprocessed with a couple of steps for further improvement. TF-IDF and word embedding were used as feature selection to perform better. All these performance measurements have been amassed to avoid under fit and overfit issue. Three encoder–decoder-based models were used against our task due to the huge potential of such kinds of models in similar tasks. For each model, the encoder part consists of convolutional neural networks that are able to effectively catch the Spatial and Temporal conditions in the text through the use of a relevant filter.

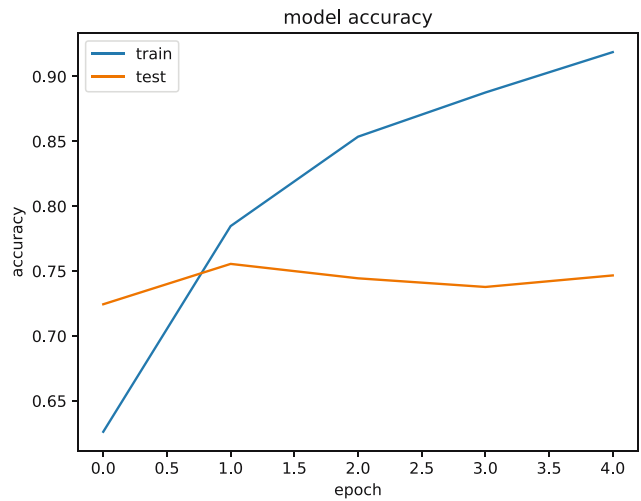

**Figure 13:** Train_test accuracy of attention model.

For the decoder part, in one of the models, LSTM was used, on the other hand, GRU was used, and finally attention mechanism was applied. Moreover, a Callback module, including the EarlyStopping function, has been used to reduce overfitting and find the best accuracy. For our multiclass classification problem, 80% training and 20% testing bring the highest accuracy. Figure 13 shows the accuracy for training and testing. Dropout for the node was kept 0.4 and recurrent dropout was kept 0.3 to prevent overfitting the model.

The final outcome is 74% accuracy for both LSTM decoder and GRU decoder models. In the attention-based decoder, 77% accurate result outperforms the previous two as shown in Table 1.

**Table 1:** Performance comparison of the proposed approach

| Name of the approach | Memory usage (MB) | Time required for training (second) | Accuracy (%) |
| --- | --- | --- | --- |
| CNN + LSTM | 432 | 193 | 74 |
| CNN + GRU | 410 | 201 | 74 |
| CNN + attention | 332 | 220 | 77 |

Here, attention-based encoder–decoder achieves the highest accuracy. The model gives the best performance with high precision (0.78), high recall (0.75), and high f1-score (0.78), while LSTM based one achieved precision 0.72, recall 0.71, f1-score 0.72 and GRU one achieved 0.70 in precision, recall and f1-score 0.69.

**Table 2:** Comparing algorithm among the binary-class labels

| Paper | Random forest | LSTM | GRU | Attention Mechanism | SVM |
| --- | --- | --- | --- | --- | --- |
| Paper [27] | 78% | – | – | – | 82% |
| Paper [28] | – | 77% | – | – | 78% |
| This paper | – | 80% | 83% | 88% | – |

In Table 2, algorithm among the binary-class labels is compared and in Table 3 algorithm among multi-class labels is compared.

**Table 3:** Comparing algorithm among multi-class labels

| Paper | SVC | Attention mechanism | Naive bayes | Random forest | GRU | LSTM |
|---|---|---|---|---|---|---|
| Paper [25] | 50% | – | 24% | 52% | 70% | – |
| This paper | – | 77% | – | – | 74% | 74% |

In the aforementioned comparisons, it can be seen that the attention mechanism performs well and better among the rest algorithms on both binary and multi-class label datasets.

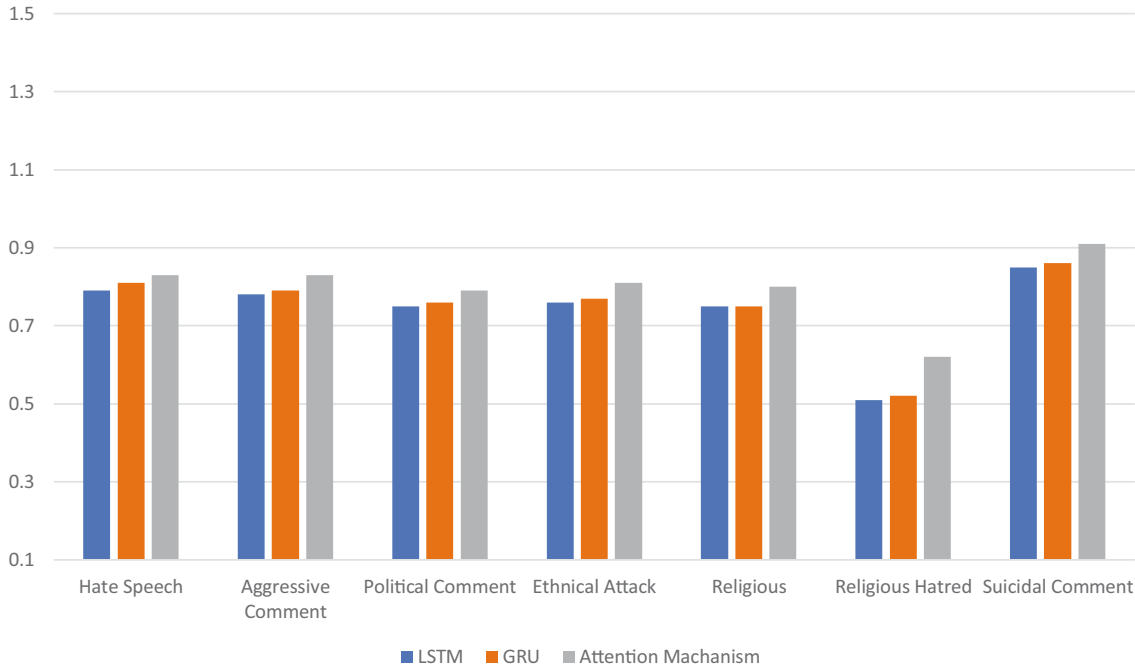

**Figure 14:** The f1-score for LSTM (Long short-term memory), GRU (Gated Recurrent Unit), and attention mechanism.

The test accuracy for various ML algorithms is shown in Figure 14 and shows the performance of the f1-score for different hate speech.

## 5 Conclusion and future work

In this article, the dataset collection methodology and annotation process were introduced to shape Bangladesh's social context. Several ML approaches were applied to our dataset, which was collected from various Facebook pages. These data include comments with seven categories: Hate Speech, Aggressive Comment, Religious Hatred, Ethnical Attack, Religious, Political Comment, and Suicidal Comment. The hateful content was divided into two sections (hateful and non-hateful), which reflects a more thorough analysis of people's patterns of use on Facebook. Finally, various ML algorithms and a deep neural network model were examined to find a suitable model and compare the algorithms' performance. The data-set can be improved for further analysis, and the outcome for future projects can be used for the benchmark.

Different Facebook pages have different types of comments filled with both hateful and non-hateful speeches. In the future, the model can be improved by creating a data parsing model for which comments will be automatically classified into their class types, whenever the link of an individual Facebook post is inserted. The categories will be presented using a graph with percentages defined. Some people who have a taste of commenting on social media with the mixing of different languages. The existing dataset can be increased with comments combined with English-Bangla from Facebook Pages and apply several ML algorithms to achieve better accuracy. Some people also have an interest in commenting on a page or post with photos or pictures rather than writing a comment. These are called photo comments. Photo comments are far easier to comment with, rather than expressing them by writing. Detecting the type of speech written on it, from the photo comments, could contribute to this model.

**Conflict of interest::** Authors state no conflict of interest.

# References

[1] Duggan M. Online harassment 2017. Washington: Pew Research Center; 2017.
[2] Emon EA, Rahman S, Banarjee J, Das AK, Mittra T. A deep learning approach to detect abusive bengali text. In 2019 7th International Conference on Smart Computing Communications (ICSCC), 2019. p. 1–5. doi: 10.1109/ICSCC.2019.8843606.
[3] Kallas P. Top 15 most popular social networking sites and apps [august 2018][online], 2018. https://www.dreamgrow.com/top-15-most-popularsocial-networking-sites/.
[4] Tarik I. Demographics of facebook population in Bangladesh, April 2018. http://digiology.xyz/demographicsfacebook-population-bangladesh-april-2018/.
[5] W. A. Social and hootsuite. (2018) 2018 digital yearbook[online]. 2018. https://digitalreport.wearesocial.com/.
[6] Unb D. 49% bangladeshi school pupils face cyberbullying[online]. 2016. https://cutt.ly/Wh7h8x0.
[7] Ethnologue. List of languages by number of native speakers, May 21 2020. https://en.wikipedia.org/wiki/List_of_languages_by_number_of_native_speakers.
[8] Islam J, Mubassira M, Islam MR, Das AK. A speech recognition system for bengali language using recurrent neural network. In 2019 IEEE 4th International Conference on Computer and Communication Systems (ICCCS); 2019. p. 73–6. doi: 10.1109/CCOMS.2019.8821629.
[9] Mumu TF, Munni IJ, Das AK. Depressed people detection from bangla social media status using lstm and cnn approach. J Eng Adv. Mar. 2021;2:41–7. doi: 10.38032/jea.2021.01.006.
[10] de Gibert O, Perez N, Garca-Pablos A, Cuadros M. Hate speech dataset from a white supremacy forum. 2018; arXiv:http://arXiv.org/abs/arXiv:1809.04444.
[11] Tuhin RA, Paul BK, Nawrine F, Akter M, Das AK. An automated system of sentiment analysis from bangla text using supervised learning techniques. In 2019 IEEE 4th International Conference on Computer and Communication Systems (ICCCS).; 2019. p. 360–4. doi: 10.1109/CCOMS.2019.8821658.
[12] Malmasi S. Detecting hate speech in social media. 2017; arXiv:http://arXiv.org/abs/arXiv:1712.06427.
[13] Rakib OF, Akter S, Khan MA, Das AK, Habibullah KM. Bangla word prediction and sentence completion using gru: An extended version of rnn on n-gram language model. In 2019 International Conference on Sustainable Technologies for Industry 4.0 (STI); 2019. p. 1–6. doi: 10.1109/STI47673.2019.9068063.
[14] Gao L, Huang R, Detecting online hate speech using context aware models. 2017; arXiv:http://arXiv.org/abs/arXiv:1710.07395.
[15] Nobata C, Tetreault J, Thomas A, Mehdad Y, Chang Y. Abusive language detection in online user content. In Proceedings of the 25th international conference on world wide web; 2016. p. 145–53.
[16] Chandrasekharan E, Samory M, Srinivasan A, Gilbert E. The bag of communities: Identifying abusive behavior online with preexisting internet data. In Proceedings of the 2017 CHI Conference on Human Factors in Computing Systems; 2017. p. 3175–87.
[17] Rodrguez A, Argueta C, Chen Y-L. Automatic detection of hate speech on facebook using sentiment and emotion analysis. In 2019 International Conference on Artificial Intelligence in Information and Communication (ICAIIC), IEEE; 2019. p. 169–74.
[18] Niam IMA, Irawan B, Setianingsih C, Putra BP. Hate speech detection using latent semantic analysis (lsa) method based on image. In 2018 International Conference on Control Electronics, Renewable Energy and Communications (ICCEREC), IEEE; 2018. p. 166–71.
[19] Nugroho K, Noersasongko E, Muljono P, Fanani AZ, Affandy, Basuki RS. Improving random forest method to detect hatespeech and offensive word. In 2019 International Conference on Information and Communications Technology (ICOIACT), IEEE; 2019. p. 514–8.
[20] Mossie Z, Wang J-H, Nagamalai D. Social network hate speech detection for amharic language. Comp Sci Inf Technol. 2018:41–55 doi: 10.5121/csit.2018.80604.

[21] Aulia N, Budi I. Hate speech detection on indonesian long text documents using machine learning approach. In Proceedings of the 2019 5th International Conference on Computing and Artificial Intelligence; 2019. p. 164–9.

[22] Mandl T, Modha S, Majumder P, Patel D, Dave M, Mandlia C, Patel A. Overview of the hasoc track at fire 2019: Hate speech and offensive content identification in indo-european languages. In Proceedings of the 11th Forum for Information Retrieval Evaluation; 2019. p. 14–7.

[23] Hossain MM, Labib MF, Rifat AS, Das AK, Mukta M. Auto-correction of english to bengali transliteration system using levenshtein distance. In 2019 7th International Conference on Smart Computing Communications (ICSCC); 2019. p. 1–5. doi: 10.1109/ICSCC.2019.8843613.

[24] Drovo MD, Chowdhury M, Uday SI, Das AK. Named entity recognition in bengali text using merged hidden markov model and rule base approach. In 2019 7th International Conference on Smart Computing Communications (ICSCC); 2019. p. 1–5. doi: 10.1109/ICSCC.2019.8843661.

[25] Md Ishmam A, Sharmin S. Hateful speech detection in public facebook pages for the bengali language 2019 18th IEEE International Conference On Machine Learning And Applications (ICMLA), IEEE; 2019. p. 555–60.

[26] Luxton DD, June JD, Fairall JM. Social media and suicide: a public health perspective. American journal of public health 2012;102(S2):S195–S200.

[27] Eshan SC, Hasan MS. An application of machine learning to detect abusive bengali text. In 2017 20th International Conference of Computer and Information Technology (ICCIT), IEEE; 2017. p. 1–6.

[28] Chakraborty P, Hanif Seddiqui Md. Puja Chakraborty and Md Hanif Seddiqui. Threat and abusive language detection on social media in bengali language. In 2019 1st International Conference on Advances in Science, Engineering and Robotics Technology (ICASERT), IEEE; 2019. p. 1–6.

[29] Duggan M. 5 facts about online harassment, OCTOBER 30, 2014. http://www.pewresearch.org/fact-tank/2014/10/30/5-facts-about-online-harassment/.

[30] Das AK, Ashrafi A, Ahmmad M. Joint cognition of both human and machine for predicting criminal punishment in judicial system. In 2019 IEEE 4th International Conference on Computer and Communication Systems (ICCCS); 2019. p. 36–40. doi: 10.1109/CCOMS.2019.8821655.

[31] Mahmud MdR, Afrin M, Razzaque MdA, Miller E, Iwashige J. A rule based bengali stemmer. In 2014 International Conference on Advances in Computing, Communications and Informatics (ICACCI), IEEE; 2014. p. 2750–6.

[32] Biswas E, Das AK. Symptom-based disease detection system in bengali using convolution neural network. In 2019 7th International Conference on Smart Computing Communications (ICSCC); 2019. p. 1–5. doi: 10.1109/ICSCC.2019.8843664.

[33] ul Haque R, Mehera P, Mridha MF, Hamid A. A complete bengali stop word detection mechanism. 2019 Joint 8th International Conference on Informatics, Electronics & Vision (ICIEV) and 2019 3rd International Conference on Imaging, Vision & Pattern Recognition (icIVPR), IEEE; 2019. p. 103–7.

[34] Emojipedia. Grinning face with smiling eyes emoji, 2014. https://cutt.ly/Jh7jte1.

[35] Subramanian D. Emoticon and emoji in text mining. 2019. https://cutt.ly/Zh7joxH.

[36] List of emoticons-Wikipedia. 2018. https://cutt.ly/lh7jlGS.

[37] Emoji. Full emoji list, v13.1. 2020. http://unicode.org/emoji/charts/full-emoji-list.html.

[38] Bahdanau D, Cho K, Bengio Y. Neural machine translation by jointly learning to align and translate. 2014. arXiv:http://arXiv.org/abs/arXiv:1409.0473.